\documentclass{article}
\usepackage[preprint]{neurips_2023}
\usepackage{natbib}
\setcitestyle{numbers,square}

\usepackage[utf8]{inputenc} % allow utf-8 input
\usepackage[T1]{fontenc}    % use 8-bit T1 fonts
\usepackage{hyperref}       % hyperlinks
\usepackage{url}            % simple URL typesetting
\usepackage{booktabs}       % professional-quality tables
\usepackage{amsfonts}       % blackboard math symbols
\usepackage{nicefrac}       % compact symbols for 1/2, etc.
\usepackage{microtype}      % microtypography
\usepackage{xcolor}         % colors

\usepackage{algorithm}
\usepackage{algorithmic}
\usepackage{threeparttable}

\usepackage{amsmath,amssymb}
\usepackage{bbding}
\usepackage{bbm}
\usepackage{subfigure}
\usepackage{graphicx}
\graphicspath{
    {./figs/}
}
\usepackage{multirow}
\usepackage{wrapfig}

\usepackage[normalem]{ulem}
\useunder{\uline}{\ul}{}
\usepackage{listings}

\newcommand{\eg}{{\em e.g.}}           % e.g.
\definecolor{light-gray}{gray}{0.6}

\title{Rethinking Data Perturbation and Model Stabilization for Semi-supervised Medical Image Segmentation}

\author{
  Zhen Zhao\\
  University of Sydney\\
  \texttt{zhen.zhao@usyd.edu.au} \\
  \And
  Ye Liu, Meng Zhao, Di Yin \\
  Tencent Youtu Lab \\
  \texttt{\{rafelliu,alexmzhao,endymecyyin\}@tencent.com}\\
  \AND
  Yixuan Yuan\\
  Chinese University of Hong Kong\\
  \texttt{yxyuan@ee.cuhk.edu.hk}\\
  \And
  Luping Zhou\\
  University of Sydney\\
  \texttt{luping.zhou@sydney.edu.au}
}

\begin{document}

\maketitle

\begin{abstract}

Studies on semi-supervised medical image segmentation (SSMIS) have seen fast progress recently. Due to the limited labelled data, SSMIS methods mainly focus on effectively leveraging unlabeled data to enhance the segmentation performance. However, despite their promising performance, current state-of-the-art methods often prioritize integrating complex techniques and loss terms rather than addressing the core challenges of semi-supervised scenarios directly. We argue that the key to SSMIS lies in generating substantial and appropriate prediction disagreement on unlabeled data.
To this end, we emphasize the crutiality of data perturbation and model stabilization in semi-supervised segmentation, and propose a simple yet effective approach to boost SSMIS performance significantly, dubbed DPMS.
Specifically, we first revisit SSMIS from three distinct perspectives: the data, the model, and the loss, and conduct a comprehensive study of corresponding strategies to examine their effectiveness.
Based on these examinations, we then propose DPMS, which adopts a plain teacher-student framework with a standard supervised loss and unsupervised consistency loss. To produce appropriate prediction disagreements, DPMS perturbs the unlabeled data via strong augmentations to enlarge prediction disagreements considerably. 
On the other hand, using EMA teacher when strong augmentation is applied does not necessarily improve performance.
DPMS further utilizes a forwarding-twice and momentum updating strategies for normalization statistics to stabilize the training on unlabeled data effectively. 
Despite its simplicity, DPMS can obtain new state-of-the-art performance on the public 2D ACDC and 3D LA datasets across various semi-supervised settings, e.g. obtaining a remarkable 22.62\% improvement against previous SOTA on ACDC with 5\% labels.
Code and logs are available at \url{https://github.com/ZhenZHAO/DPMS}. 

\end{abstract}

%%%%%%%%%%%%%%%%%%%%%%%%%%%%%%%%%%%
%%  1. introduction
%%%%%%%%%%%%%%%%%%%%%%%%%%%%%%%%%%%%
% \clearpage

\section{Introduction}

\begin{figure*}
  \centering
  \subfigure[Performance Comparisons]{
  \centering
   \includegraphics[width=0.48\linewidth]{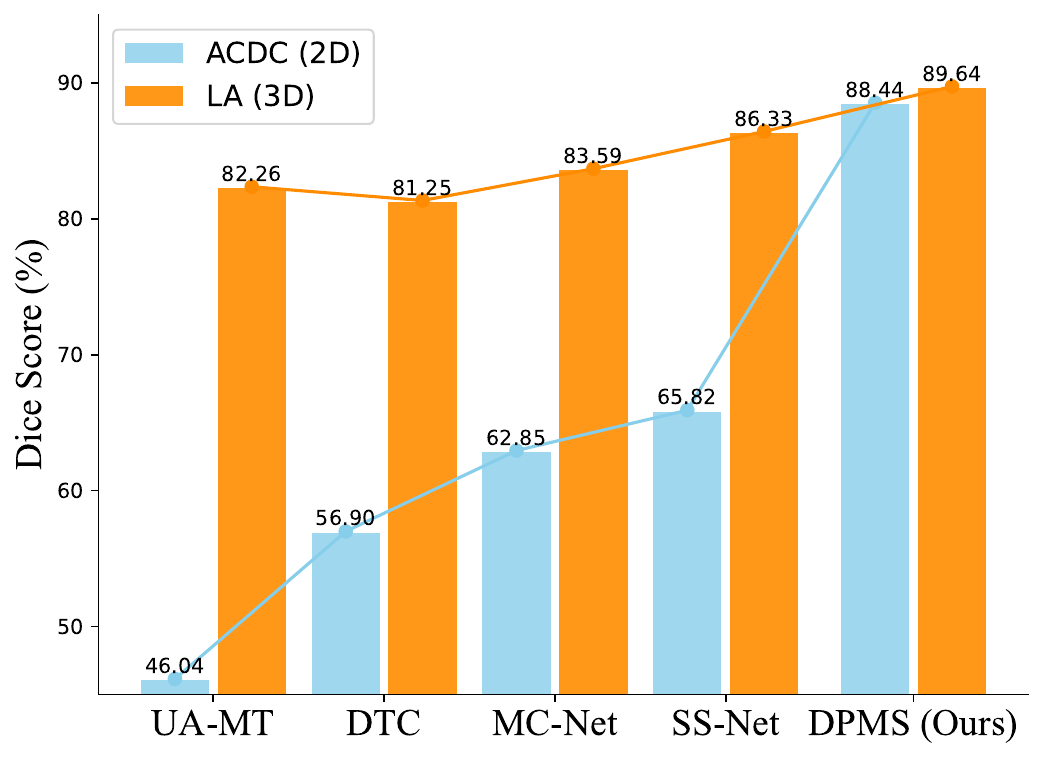}
   \label{fig:DPMS:performance}
  }
  \hfill
  \subfigure[Diagram of our proposed DPMS]{
  \centering
   \includegraphics[width=0.45\linewidth]{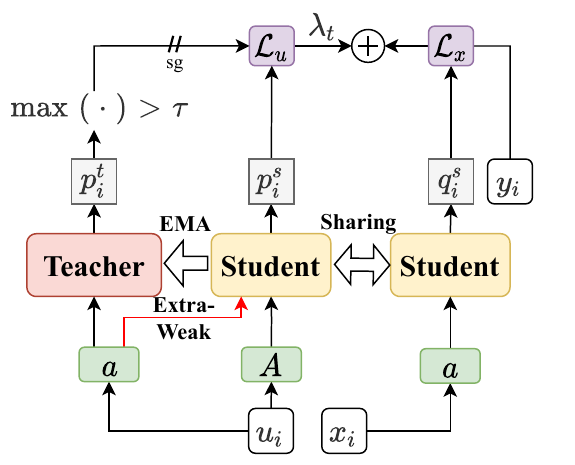}
   \label{fig:DPMS:diagram}
  }
  \caption{(a) We compare our DPMS with recent SSIMS methods in terms of the Dice score (\%) on 2D ACDC and 3d LA datasets with only 5\% labeled data. Remarkable performance gains can be observed. (b) DPMS employs weak and strong augmentations (denoted by $a$ and $A$, respectively) to perturb unlabeled inputs $u_i$. In a standard teacher-student framework, the student model is trained on the provided labeled data $(x_i, y_i)$ via a standard supervised loss $\mathcal{L}_x$, as well as on the unlabeled data $u_i$ via a consistency loss $\mathcal{L}_u$ supervised by pseudo-labels generated from the teacher model. The weights and buffer statistics of the teacher model are updated using the exponential moving average of the corresponding values from the student model. Pseudo-labels $p_i^t$ are further filtered by a high-confidence threshold $\tau$. A ramp-up term $\lambda_t$ is adopted to leverage the unlabeled data gradually.
  }
  \label{fig:DPMS:intro}
\end{figure*}
Medical image segmentation is an urgent vision task that contributes to medical image reasoning, which is vital in the development of the computer-aided diagnosis (CAD) system~\cite{valanarasu2022unext, wu2022mutual, ji2022amos, lin2022ds}. 
Conventional supervised medical image segmentation methods rely heavily on extensive pixel-level annotated data for the model training. 
In practical medical applications, obtaining large amounts of fine-grained annotated data is costly and even infeasible, greatly hindering their wide applications~\cite{you2022momentum, you2022simcvd}. To this end, many studies have been focused on semi-supervised medical image segmentation (SSMIS), which aims at learning a deep segmentation model by using a limited number of annotated medical images and abundant unlabeled medical images to achieve satisfied segmentation performance~\cite{sss22uamt, wu2022mutual, bai2023bidirectional, wu2022exploring, luo2022semi}.

Following the research line of semi-supervised semantic segmentation on natural images~\cite{sss22st++, sss22PSMT}, SSMIS studies have evolved from earlier self-training based methods~\cite{bai2017semi,wang2021deep} into recent dominant consistency regularization (CR) based approaches~\cite{sss22uamt,wu2022exploring}. CR-based approached methods, like Mean-Teacher~\cite{ssl17mt} and FixMatch~\cite{semifix}, leverage the label-preserving data or model perturbations to encourage prediction consistency on differently perturbed views from the same input. The key to such methods is to generate prediction disagreements on unlabeled data~\cite{zhao2022augmentation}.  To further improve SSMIS performance, recent studies tend to introduce more advanced and complicated techniques, such as additional transformer-based branch~\cite{luo2022semi}, extra feature-level perturbations and constraints~\cite{wu2022exploring}, additional self-supervised contrastive losses~\cite{peng2021self}.
Despite their impressive performance, these methods usually  come at the cost of increasingly complex designs and benefit SSMIS in an indirect manner. 
% Considering the fact that the key to SSMIS lies in generating appropriate prediction disagreement on unlabeled data, 
% Instead of introducing extra components or losses, 
Differently, we focus on the semi-supervised problem itself to produce appropriate prediction disagreement, and strive to propose a simple yet highly effective method to boost SSMIS directly.

In an effort to simplify SSMIS studies, in this paper,  we diverges from complex designs and redirects the focus towards the intrinsic nature of the semi-supervised problem, \textit{i.e.,} to produce appropriate prediction disagreement.
As shown in Figure~\ref{fig:DPMS:explore}, we first conduct a thorough analysis by revisiting SSMIS from three distinct perspectives: the data, the model, and the loss supervisions.
Through a comprehensive study of their corresponding strategies, we rigorously evaluate their effectiveness and implications.
Specifically, employing data augmentations is the most straightforward and effective way to generate label-preserving disagreement in SSMIS. 
However, most of previous works focus on investigating the effectiveness of data perturbations in the natural image domain, while few works studied the effectiveness on medical images.
For example, the effectiveness and reliability of strong augmentation on medical images have not been paid sufficient attention.
Hence, we first revisit the contributions of different data perturbations to the SSMIS problem. Based on our findings, we figure out that data perturbations can yield sufficient prediction disagreements and boost the SSMIS performance. However, as discussed in \cite{sss21simple,zhao2022augmentation}, too strong data perturbations will inevitably hurt the distribution of the original data and consequently degrade the performance. To tackle the issue, we further study the model stabilization in SSMIS, where we come up with two simple yet effective model stabilization strategies, \textit{i.e.}, the EMA-BN and Extra-Weak, to prevent the model statistics from being severely disturbed. As a result, we highlight the significance of data perturbation and model stabilization in SSMIS.

% The extensive data perturbations aim to yield sufficient prediction disagreements while the model stabilization can prevent performance degradation derived from data over-perturbations~\cite{sss21simple,zhao2022augmentation}.

Motivated by our revisiting, we propose DPMS, that adopts a simple teacher-student framework to employ the effective \textbf{D}ata \textbf{P}erturbation and \textbf{M}odel \textbf{S}tabilization strategies to boost SSMIS. 
On the one hand, DPMS poisons unlabeled instances via various strong augmentations, including geometrical transformations, intensity-based perturbation and copy-paste, to enlarge prediction disagreements considerably. On the other hand, DPMS utilizes an extra forward (forwarding unlabeled data into the student model) and momentum updating strategies for normalization statistics to stabilize the training on unlabeled data effectively. Without bells and whistles, our simple DPMS can achieve remarkable performance improvement compared to current state-of-the-art (SOTA) SSMIS methods. As shown in Figure~\ref{fig:DPMS:performance}, DPMS consistently outperform other methods by a large margin, \textbf{e.g.} obtaining a remarkable 22.62\% Dice improvement compared to previous SOTA SS-Net on ACDC with 3 labeled samples. Our main contributions are summarized as follows.

\begin{itemize}
    \item We revisit the key elements of data, model, and loss supervision in semi-supervised medical image segmentation. Through in-depth analysis, we conduct comprehensive studies on various strategies associated with each element.
    \item We break the trend of recent SSMIS studies that tend to introduce increasingly complicated designs and propose a simple yet effective DPMS that emphasize the significance of data perturbation and model stabilization to boost SSMIS.
    \item Benefiting from the perturbing and stabilizing designs, our simple DPMS can readily achieve new SOTA performance on public 2D and 3D SSMIS benchmarks.
\end{itemize}

% Benefiting from such mutual restraining yet promoting designs,

% Our work is based on simple techniques, supported by favourable empirical
% results, and validated by extensive ablation experiments, at our best.

% \clearpage
%%%%%%%%%%%%%%%%%%%%%%%%%%%%%%%%%%%%
%%  2. Related works
%%%%%%%%%%%%%%%%%%%%%%%%%%%%%%%%%%%%
\section{Related work}
\label{sec:rwork}

\subsection{Supervised semantic segmenation}

Deep learning-based techniques have exhibited remarkable effectiveness in the realm of natural image segmentation, such as the DeepLab series~\cite{chen2014semantic, seg17deeplab, chen2017rethinking, seg18deeplabv3plus}, the HRNet~\cite{wang2020deep}, the PSPNet~\cite{zhao2017pyramid}, \emph{etc}. Most of them  leverage a pre-trained ResNet~\cite{he16resnet} as the backbone encoder to extract semantic information and employ diverse decoders to generate dense predictions. Differently, medical images have some special properties like scarce labeled data, and fine-grained classes, and smooth boundaries. To address these challenges, numerous studies have devised specialized methods for medical image segmentation, which can be classified into two primary categories. The first focuses mainly on designing medical-specific network architecture, like the widely used UNet~\cite{seg15unet} and Vnet~\cite{milletari2016v}, which designs a fully convolutional network that is trained end-to-end to capture multi-level semantic features to perform dense predictions. The second tends to propose medical-specific loss functions, like the Dice loss~\cite{milletari2016v}, which utilizes the dice coefficient to tackle the class imbalance problem. Nevertheless, the majority of these approaches heavily depend on extensively annotated medical image datasets, which necessitates significant effort from expert annotators to obtain precise annotations~\cite{wang2020noise}.
% ~\cite{wang2020noise, guo2022joint}.

\subsection{Semi-supervised semantic segmentation}
To alleviate the challenges associated with labor-intensive dense annotations, semi-supervised segmentation (SSS) has emerged as a promising solution in addressing labeling difficulties~\cite{sss18advseg, sss19s3gan, ssl20bias, sss21c3seg}. Current SSS methods in the natural image domain can be divided into two main categories, \emph{i.e.}, the self-training (ST)~\cite{ssl13pseudo,ssl22lassl} based methods and the consistency regularization (CR) based methods~\cite{zhao2022instance, ssl22dcssl}. The former approaches~\cite{sss22st++, sss21cps} focus on pseudo-labeled sample selection for fine-tuning the model while CR based methods~\cite{sss22PSMT, sss22u2pl} aim at enabling the model to generate consistent predictions on differently perturbed views from the same instance. CR-based methods employ various data-level and model-level perturbations to obtain state-of-the-art SSS performance.

Currently, most SSS methods in the medical image domain follow the same designing ideas as natural images~\cite{sss22uamt, wu2022mutual, bai2023bidirectional, wu2022exploring, luo2022semi}.
UA-MT~\cite{sss22uamt} uses a mean-teacher framework and encourages the student model to gradually generate consistent predictions as the teacher model based on the proposed uncertainty-aware training scheme. 
SASSNet~\cite{li2020shape} further enforces a geometric shape constraint upon the segmentation outputs.
DTC~\cite{luo2021semi} designs an additional task-level constraint into a dual task-consistency framework. Recent state-of-the-art methods tend to introduce more advance techniques to further improve SSMIS performance.
MC-Net~\cite{wu2022mutual} perturbs the predictions with multiple different decoders and encourages the prediction consistency between the perturbed decoders. 
Authors in~\cite{peng2021self} propose to pre-train the image encoder with meta-labels and then introduce an extra self-paced contrastive learning in semi-supervised framework.
CT-CT~\cite{luo2022semi} introduces an extra transformer branch and encourages prediction consistency between the CNN model and the Transformer model to enable the model to benefit from the two learning paradigms.
SS-Net~\cite{wu2022exploring} employs the feature-level virtual adversarial training (VAT) and prototype-level scattering losses to achieve promising performance.
% BCP~\cite{bai2023bidirectional} joint processes the labeled and unlabeled data using a copy-paste strategy and encourages the prediction consistency between the teacher model and the student model.
Despite their impressive performance, we clearly observe that SSMIS studies along this line come at the cost of introducing 
more complex techniques, e.g., extra network structures or
additional training procedures and losses. Differently, in this work, we redirects the focus towards the semi-supervised problem itself, and highlight the cruciality of data perturbation and model stabilization to generate substantial and appropriate prediction disagreement in SSMIS.

\begin{figure}[t]
    \centering
    \includegraphics[width=0.99\linewidth]{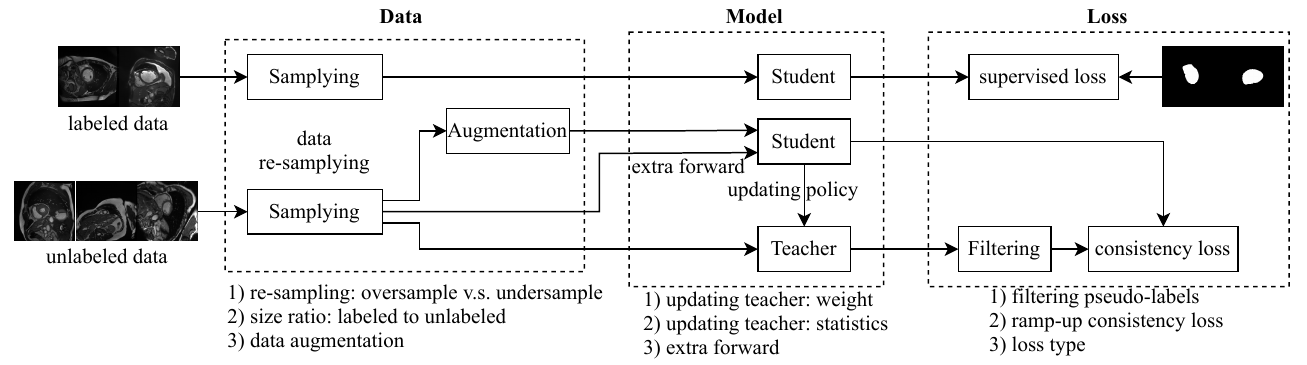}
    \caption{Revisiting SSMIS in terms of the data, the model and the loss supervision.}
    \label{fig:DPMS:explore}
\end{figure}

% \clearpage
%%%%%%%%%%%%%%%%%%%%%%%%%%%%%%%%%%%%
%%  3. Methods
%%%%%%%%%%%%%%%%%%%%%%%%%%%%%%%%%%%%

\section{Method}
\label{sec:method}

In this section, we first introduce the formulation of SSMIS in Sec.~\ref{method:formulation} and then provide a comprehensive revisiting on core elements of the data, the model and the loss supervision in SSMIS in Sec.~\ref{method:revisit}. Finally, Sec.~\ref{method:dpms} describe our proposed DPMS in detail.

\subsection{Problem Formulation}
\label{method:formulation}

In a common SSMIS task, labeled data $\mathcal{X}$ and unlabeled data $\mathcal{U}$ are provided, with typically $|X| \ll |U|$. In terms of the training process, let $\mathcal{B}_x= \{(x_i, y_i)\}_{i=1}^{B}$ be a batch of labeled samples and $\mathcal{B}_u=\{u_i\}_{i=1}^{\mu B}$ be a batch of unlabeled samples, where $\mu$ denotes the size ratio of $|\mathcal{B}_u|$ to $|\mathcal{B}_x|$. Then the goal of SSMIS is to train a deep segmentation model on both labeled an unlabeled data.

In a common teacher-student framework, the student model, parameterized by $\theta_s$, is first trained on the labeled data via a standard supervised loss $\mathcal{L}_x$, 
\begin{align}
    \mathcal{L}_x = \frac{1}{|\mathcal{B}_x|} \sum_{i=1}^{B} \frac{1}{H\times W}\sum_{j=1}^{H\times W} \ell(\hat{y}_i(j), y_i(j)),
\end{align}
where $\hat{y}_i$ denotes the student model's prediction output on the weakly augmented input $x_i$, \textit{i.e.,} $\hat{y}_i\!=\!f(a(x_i); \theta_s)$, and $j$ represents the $j$-th pixel on the image or the corresponding segmentation mask with a resolution of $H \times W$. $\ell$ represents the loss function used to supervise the training, which can be dice loss, cross-entropy loss, or a compound loss of both. Weak augmentations, denoted by $a(\cdot)$, include random geometrical transformations, like random cropping and flipping operations. The teacher model, parameterized by $\theta_t$, is typically not trained directly on the labeled or unlabeled data, but updated by the weights and statistics from the student model. We discuss more in Sec~\ref{method:revisit:model}.

On the other hand, the unsupervised consistency loss on unlabeled data, denoted by $\mathcal{L_u}$, can differ from method to method. Following a standard CR-based method, the unlabeled data can be leveraged via enforcing prediction consistency on differently augmented views of the same input. Let $A(\cdot)$ and $a(\cdot)$ represent two different augmentation strategies, $\mathcal{L}_u$ can be formulated as,
\begin{align}
    \mathcal{L}_{u} = \frac{1}{|\mathcal{B}_u|}\!\sum_{i=1}^{\mu B}\!\frac{1}{H\times W}\sum_{j=1}^{H\!\times\!W}\! \psi(u_i) \, \ell(f(A(u_i);\theta_s), f(a(u_i);\theta_t)) 
\end{align}
where $\psi(u_i)$ represent the selection strategy to filter out unlabeled data with less confident predictions. In summary, the total training loss can be,
\begin{align}
    \mathcal{L} = \mathcal{L}_x + \lambda_t \mathcal{L}_u
\end{align}
where $\lambda_t$ denote the loss weight to adjust the importance of consistency loss $\mathcal{L}_u$ and can also be a  function of the iteration index $t$, \textit{i.e.,} iteration dependent.

% \newpage

\subsection{Revisiting SSMIS}
\label{method:revisit}

As shown in Figure~\ref{fig:DPMS:explore}, we then revisit the core elements of the data, the model and the loss supervisions in SSMIS studies. Extensive exploration and examination are conducted.

\subsubsection{Data: sampling and augmentation}
\label{method:revisit:data}

\begin{figure*}
  \centering
  \subfigure[Re-sampling]{
  \centering
   \includegraphics[width=0.45\linewidth]{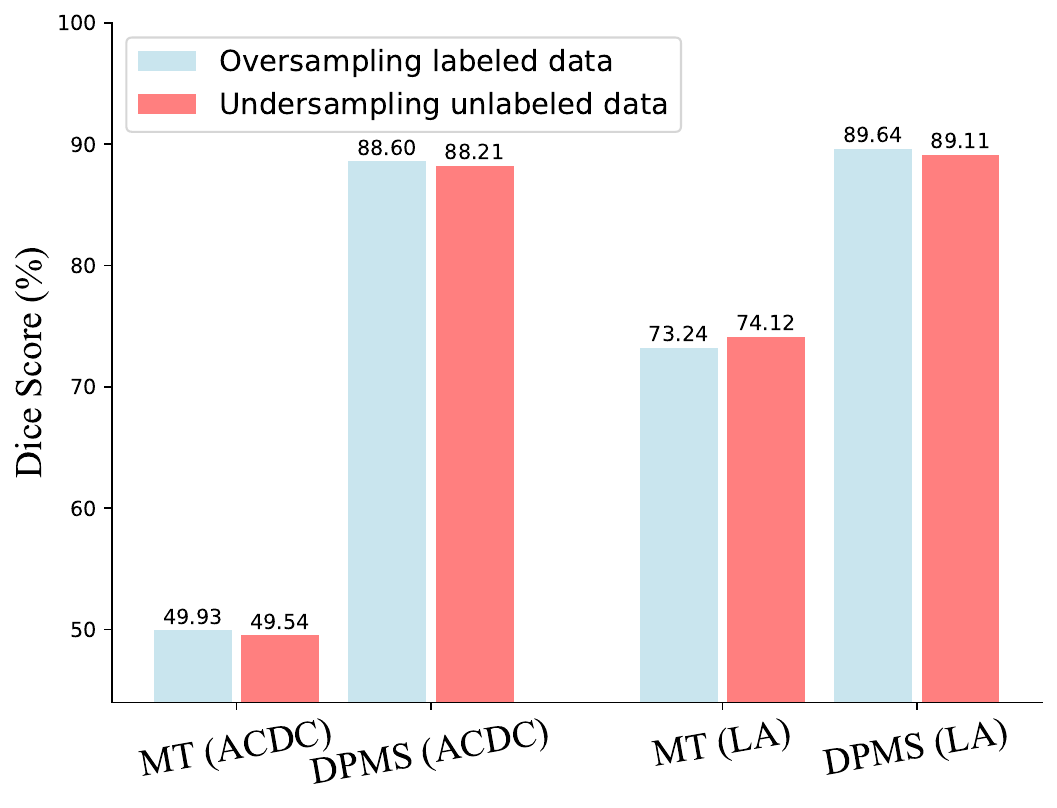}
   \label{fig:DPMS:data:sample}
  }
  \hfill
  \subfigure[Size ratios]{
  \centering
   \includegraphics[width=0.45\linewidth]{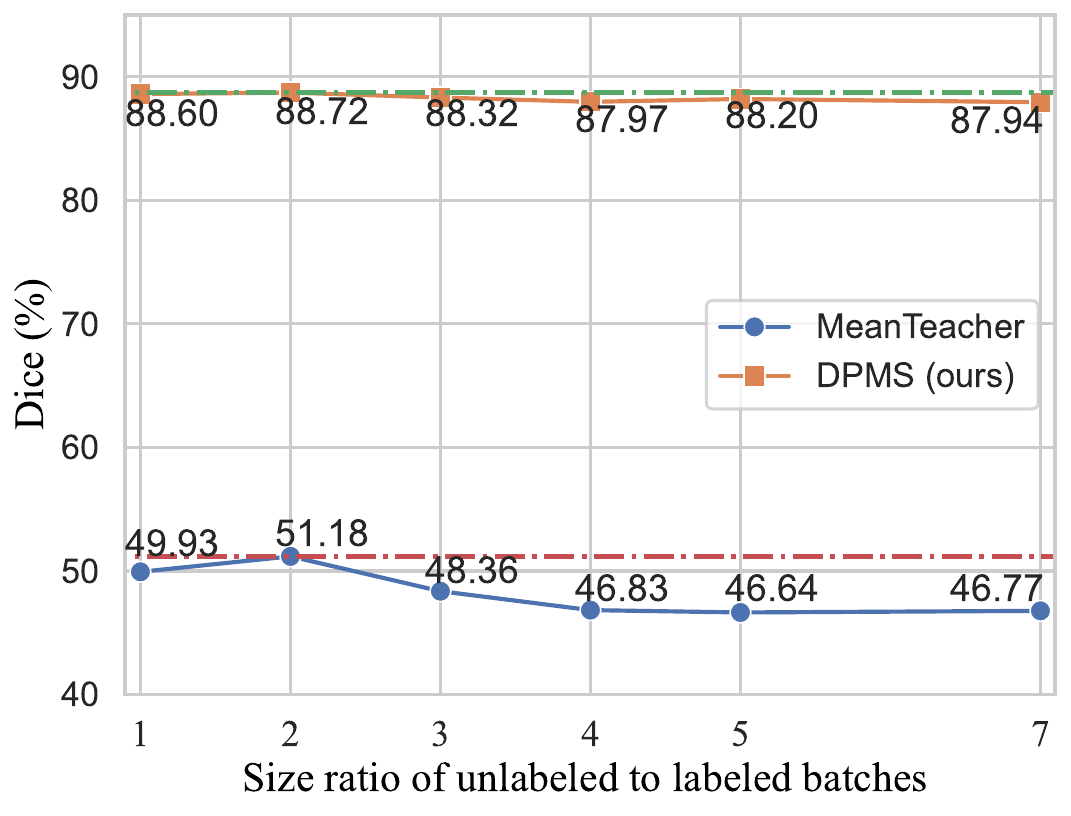}
   \label{fig:DPMS:data:ratio}
  }
  \caption{Effect of different re-sampling strategies and size ratios of unlabeled to labeled batches. By default, we apply the ``oversampling labeled data" and set the size ratio $\mu =1$ for fair comparisons.}
  \label{fig:dpms:data}
\end{figure*}

Regarding the data problem in SSMIS, we aim to address three primary questions. First, data re-sampling. Since the amount of labeled and unlabeled data differs significantly, \textit{e.g.}, $|X| \ll |U|$, employing sampling strategies is necessary to facilitate training on both labeled and unlabeled sets. Naturally, there are two different ways: 1) oversampling the labeled data and 2) under-sampling the unlabeled data. We examine the two different sampling strategies for 2D ACDC and 3D LA semi-supervised segmentations on the plain Mean-teacher and our proposed DPMS methods. As depicted in Figure~\ref{fig:DPMS:data:sample}, both re-sampling strategies demonstrate comparable and closely aligned segmentation performance across different datasets and SSMIS methods. This is simply because all the labeled and unlabeled data can be sufficiently traversed as long as the total training iterations are significant. Therefore, different re-sampling approaches will not large affect the SSMIS performance, and we over-sample the labeled data by default in our study.

Second, the size ratio $\mu$. The effect of different $\mu$ is extensively discussed in semi-supervised classification~\cite{semifix}, but rarely explored in SSMIS. Similar to the loss weight of consistency loss $\mathcal{L}_u$, larger $\mu$ will prioritize the importance of unlabeled training. We investigate its effect on ACDC datasets with different SSMIS methods. As shown in Figure~\ref{fig:DPMS:data:ratio}, we can clearly see that $\mu=2$ can achieve the best performance. However, different values of $\mu$ have less influences on our proposed DPMS. Thus we select $\mu =1$ by default for fair comparison with other SSMIS methods.

\begin{figure}[t]
  \centering
  \subfigure[Original]{
  \centering
   \includegraphics[width=0.232\linewidth]{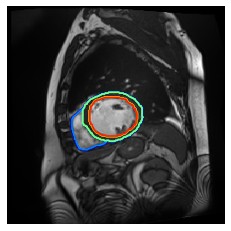}
   \label{fig:augs:org}
  }
  \hfill
  \subfigure[Geometrical Aug]{
  \centering
   \includegraphics[width=0.232\linewidth]{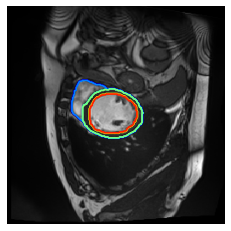}
   \label{fig:augs:geo}
  }
  \hfill
  \subfigure[Intensity-based Aug]{
  \centering
   \includegraphics[width=0.232\linewidth]{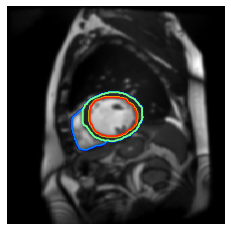}
   \label{fig:augs:col}
  }
  \hfill
  \subfigure[Copy-paste]{
  \centering
   \includegraphics[width=0.232\linewidth]{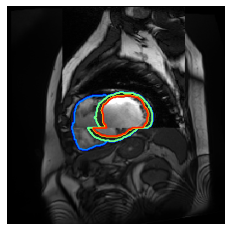}
   \label{fig:augs:mix}
  }
  \caption{Visual examples of different data augmentations on cardiac images. }
  \label{fig:augs}
\end{figure}

Third, the critical data augmentations. The key to SSMIS lies in producing appropriate prediction disagreement, and applying data augmentations can be the most straightforward and effective way to generate such label-preserving disagreement. As shown in Figure~\ref{fig:augs}, we investigate three popular kinds of data augmentations, \textit{i.e.}, geometrical transformations (random cropping and flipping),  intensity-based augmentations (randomly adjusting brightness and contrast), and Copy-and-paste~\cite{augs21copy} (widely applied in semi-supervised semantic segmentation~\cite{sss22PSMT,sss21cps,zhao2022augmentation}). 
In Table~\ref{tab:dpms:augs}, we examine the effective of each type of augmentations and their combinations on 2D ACDC and 3D LA datasets. Particularly, since the geometrical transformations will alter the image and corresponding segmentation mask, we first need to apply the same geometrical transformation when examining the intensity-based or copy-and-paste augmentations. As we expected, these strong augmentations can significant boost the SSMIS performance on both 2D and 3D datasets. Especially, the intensity-based augmentation has proven to be the most effective way to perturb the unlabeled instance in SSMIS.

{
\begin{table}[t]
\centering
% \resizebox{0.9\linewidth}{!}{
\begin{tabular}{ccc|cc}
\toprule
  \multicolumn{3}{c}{Augmentations $A(\cdot)$} & \multicolumn{2}{c}{Dice(\%)} \\ 
  \cline{1-3} \cline{4-5}
Geometrical & Intensity & Copy-paste &  ACDC (3) & LA (4)\\
\hline
\checkmark &   &   & 44.87  &  72.09 \\
\checkmark & \checkmark  &   & 64.61  & 82.07  \\
\checkmark &   &  \checkmark & 59.82  &  74.31 \\
\checkmark & \checkmark  &  \checkmark & 73.33  &  86.17 \\
\bottomrule
\end{tabular}
% }
\caption{Effect of different data augmentations on SSMIS. All the results are examined without any model stabilization strategies and with loss weight $\lambda_u=1.0$, thresholds $\tau=0.95$ and $\tau=0.8$, loss type ``Dice" and ``CE" for datasets ACDC and LA, respectively.}
\label{tab:dpms:augs}
\end{table}
}

\subsubsection{Model: updating and stabilization}
\label{method:revisit:model}

{
\begin{table}[t]
\centering
% \resizebox{0.9\linewidth}{!}{
\begin{tabular}{ccc|c|cc}
\toprule
  \multicolumn{3}{c|}{Model} & \multirow{2}{*}{Augs}& \multicolumn{2}{c}{Dice(\%)} \\ 
  \cline{1-3}  \cline{5-6}
Ema-teacher & Ema-BN & Extra-Weak &  & ACDC (3) & LA (4)\\
\hline
 &   &   &   &  45.80 & 69.59\\
\checkmark &   &   &   &  44.87 & 72.09\\
\checkmark & \checkmark  &   &   &  46.62   &76.05\\
\checkmark &   &  \checkmark &   &  48.75  &79.53\\
\checkmark & \checkmark  &  \checkmark &   &  49.24  & 79.90\\
\hline
 &   &   & \checkmark  &  80.03 & 84.46\\
\checkmark &   &   & \checkmark  &  73.33 & 86.17\\
\checkmark & \checkmark  &   & \checkmark  &  84.09   &87.83\\
\checkmark &   &  \checkmark &  \checkmark &  87.11  &88.02\\
\checkmark & \checkmark  &  \checkmark & \checkmark  &  88.44  & 89.33\\
\bottomrule
\end{tabular}
% }
\caption{
Effect of different model stabilization strategies on SSMIS. The ``Augs" denotes all kinds of augmentations in Table~\ref{tab:dpms:augs} are applied. Loss weights and thresholds are the same as in Table~\ref{tab:dpms:augs}.}
\label{tab:dpms:models}
\end{table}
}

% When considering the model design in SSMIS, two main questions arise:  1) how the pseudo-labels are generated, and 2) how the model are stabilized to prevent collapsing caused by strong data augmentations. In the literature, there are two basic ways to produce pseudo-labels. First, utilize the ensemble model, \textit{i.e.}, the teacher model, to generate pseudo-labels for unlabeled instances, like the widely applied Mean-teacher framework~\cite{semimt}. It is worth noting that the consistency-based methods are essentially the same as pseudo-labeling, where the predictions from one view serve as the pseudo-labels for another view. Second, utilize the student model itself to generate pseudo-labels, like the FixMatch~\cite{semifix}. As we can see from Table~\ref{tab:dpms:models}, 
% compared to using the student model, adopting the teacher model to generate pseudo-labels performs better on 3D LA dataset but worse on 2D ACDC dataset when applying strong data augmentation discussed in Sec.~\ref{method:revisit:data}. 
% As a result, it remains inconclusive as to which strategy is ultimately superior. In our study, we adopt the standard mean-teacher framework as our baseline, and design more stabilization strategies to further improve the performance.

When considering the model design in SSMIS, two main questions arise:  1) how the pseudo-labels are generated, and 2) how the model are stabilized to prevent collapsing caused by strong data augmentations. In the literature, there are two basic ways to produce pseudo-labels. First, utilizing the ensemble model, \textit{i.e.}, the teacher model, generates pseudo-labels for unlabeled instances, like the widely applied Mean-teacher framework~\cite{semimt}. It is worth noting that the consistency-based methods are essentially the same as pseudo-labeling, where the predictions from one view serve as the pseudo-labels for another view. Second, utilizing the student model itself generates pseudo-labels, like the FixMatch~\cite{semifix}. As we can see from Table~\ref{tab:dpms:models}, 
compared to using the student model, adopting the teacher model to generate pseudo-labels performs better on 3D LA dataset but worse on 2D ACDC dataset when applying strong data augmentation discussed in Sec.~\ref{method:revisit:data}. 
As a result, it remains inconclusive as to which strategy is ultimately superior. In our study, we adopt the standard mean-teacher framework as our baseline, and design more stabilization strategies to further improve the performance.

As elucidated in \cite{sss21simple,zhao2022augmentation}, applying strong data augmentations carries the potential risk of over-perturbations, which can hurt the data distribution and consequently degrade the SSMIS performance. Hence, ensuring the stabilization of the model becomes crucial when employing strong data perturbations in the context of semi-supervised learning. Unlike existing studies, we did not revise the augmentation strategies~\cite{zhao2022augmentation}, nor did we consider rectifying strategies like distribution-specific BN~\cite{chang2019domain}. Differently, we design two simple yet effective strategies to stabilize the training. 

\textbf{First}, in addition to updating the weights of the teacher model, we also updating the batch normalization (BN) statistics via the exponential moving average (EMA) of BN of the student model.
\begin{align}
    \theta_t & \leftarrow \alpha \theta_t + (1 - \alpha) \theta_s, \label{equ:teacher:weight}\\
    \nu_t & \leftarrow \alpha \nu_t + (1 - \alpha) \nu_s, \label{equ:teacher:bn}
\end{align}
where $\nu_t$ and $\nu_s$ represent the BN statistics of the teacher and the student model, respectively. $\alpha$ is a momentum parameter, set as 0.99 by default. We can see from Table~\ref{tab:dpms:models} that applying EMA-BN can effectively improve the SSMIS performance on both 2D ACDC and 3D LA datasets. Improvements become even more pronounced when strong data augmentation is applied. For instance, using EMA-BN yields an improvement of 10.76\% compared to the augmentation baseline using EMA-teacher on ACDC, which further emphasizes the importance of model stabilization.

\textbf{Second}, to further stabilize the BN statistics, we forward the weakly augmented inputs to the student model, not only the teacher model, dubbed as ``Extra-weak". Despite its embarrassing simplicity, we can see from Table~\ref{tab:dpms:models} that ``Extra-weak" can effectively boost the SSMIS performance, especially when the strong data augmentations are applied. Indeed, applying all these stabilization strategies can successfully improve the augmentation baseline by a large margin on both 2D and 3D SSMIS tasks.

\subsubsection{Loss supervision: loss type and filtering}

{
\begin{table}[t]
\centering
% \resizebox{0.9\linewidth}{!}{
\begin{tabular}{ccc|c}
\toprule
  \multicolumn{3}{c}{Loss} & Dice(\%) \\ 
  \cline{1-3} \cline{4-4} 
loss type & Threshold & Ramp-up &  ACDC (3)\\
\hline
``dice+ce" & \checkmark & \checkmark  &  88.07    \\
``ce" &  \checkmark &  \checkmark &  87.05    \\
``dice" &  \checkmark &  \checkmark &  \textbf{88.44}    \\
\hline
``dice" &   &   & 86.53     \\
``dice" &  \checkmark &   &  88.22    \\
``dice" &   & \checkmark  & 87.25     \\
\bottomrule
\end{tabular}
% }
\caption{
Effect of different consistency loss supervisions on 2d ACDC dataset with 3 labeled samples.}
\label{tab:dpms:loss:acdc}
\end{table}
}

{
\begin{table}[t]
\centering
% \resizebox{0.9\linewidth}{!}{
\begin{tabular}{ccc|c}
\toprule
  \multicolumn{3}{c}{Loss} & Dice(\%) \\ 
  \cline{1-3} \cline{4-4} 
loss type & Threshold & Ramp-up &  LA (4)\\
\hline
``dice+ce" & \checkmark & \checkmark  &  89.02    \\
``ce" &  \checkmark &  \checkmark &  \textbf{89.33}    \\
``dice" &  \checkmark &  \checkmark &  88.82    \\
\hline
``ce" &   &   & 83.68     \\
``ce" &  \checkmark &   &  87.78    \\
``ce" &   & \checkmark  & 88.30    \\
\bottomrule
\end{tabular}
% }
\caption{
Effect of different consistency loss supervisions on 3d LA dataset with 4 labeled samples.}
\label{tab:dpms:loss:la}
\end{table}
}

In terms of the loss supervision in SSMIS, we explore following three questions: 1) the appropriate loss type, 2) pseudo-label selections, and 3) the ramp-up policy of unlabeled consistency loss. In the literature, there are three widely adopted losses, the Dice loss, the cross-entropy (CE) loss, and the compound loss of both. As we can see from Tables~\ref{tab:dpms:loss:acdc} and \ref{tab:dpms:loss:la}, different loss types can achieve close and comparable performance. In specific, the ``dice loss" has demonstrated superior performance  on 2D ACDC, while the ``ce loss" has shown optimal results for 3D LA. Considering that the ACDC dataset comprises four distinct classes while the LA dataset contains only two classes, the Dice loss emerges as the more appropriate option for ACDC due to its capability to alleviate class imbalance issues.

Employing pseudo-label selection process has been widely studied in SSMIS. It is regarded as an effective and necessary procedure to address the confirmation bias or accumulated errors in semi-supervised learning~\cite{bias2019}. Following the FixMatch~\cite{semifix}, we simply adopt a pre-defined threshold, denoted as $\tau$, to filter out the unlabeled data with less confident pseudo-labels. As we can see from Tables~\ref{tab:dpms:loss:acdc} and \ref{tab:dpms:loss:la}, a high-confidence threshold can effectively improve the SSMIS performance, \textit{e.g.,} yielding 4.1\% and 1.69\% Dice improvements on LA and ACDC, respectively. More detailed ablations studies on the threshold $\tau$ is provided in Sec.~\ref{exp:abls}.

Following Mean-Teacher~\cite{semimt} and UA-MT~\cite{sss22uamt}, we further investigate the effect of the ramp-up policy of the unsupervised loss in SSMIS. As discussed in PI-model~\cite{semipi}, we adopt a simple iteration-dependent loss weight ramp-up function $\lambda_t$ to release the impact of consistency loss gradually. Specifically, $\lambda_t$ is starting from zero and ramping up along a Gaussian curve during the first 150 epochs to the ultimate value of $\lambda_u$. The ablation study of the hyper-parameter $\lambda_u$ is provided in Sec.~\ref{exp:abls}. As shown Table~\ref{tab:dpms:loss:la}, applying ramp-up strategy can bring a remarkable performance improvement of 4.62\% on LA dataset with 4 labeled samples.

\subsection{Our method: DPMS}
\label{method:dpms}

Based on the above comprehensive revisiting, we can clearly observe the significance of the data perturbation and model stabilization in SSMIS. To this end, instead of integrating other complicated designs like contrastive losses~\cite{peng2021self}, additional transformer branches~\cite{luo2022semi}, we simply follow a plain mean-teacher framework, and propose our method DPMS by integrating our explored effective perturbing and stabilizing strategies. As shown in Figure~\ref{fig:DPMS:diagram}, we first employ the weak and strong data augmentations, $a(\cdot)$ and $A(\cdot)$ on data inputs, and feed the augmented data into the student and teacher model to obtain predictions accordingly,
\begin{align}
    p^s_i &= f (A(u_i); \theta_s), \\
    q^s_i &= f (a(x_i); \theta_s), \\
    p^t_i &= f (a(u_i)); \theta_t).
\end{align}
Then the student model can be trained on both labeled and unlabeled data by the total loss,
\begin{align}
    \mathcal{L} = \frac{1}{B} \sum_{i=1}^{B} \frac{1}{H\times W}\sum_{j=1}^{H\times W} \ell(q^s_i(j), y_i(j)) + \lambda_t \,\mathbbm{1}(\max(p^t_i(j))\!\geq\!\tau) \ell(p^s_i(j), p^t_i(j)),
\end{align}
where $\mathbbm{1}(\max(p^t_i(j))\!\geq\!\tau)$ represent to retain the pseudo-labels whose maximum probability is higher than the pre-defined high-confidence threshold $\tau$. On the other hand, the teacher model is trained by using Equations~\ref{equ:teacher:weight} and \ref{equ:teacher:bn} to update its weights and BN statistics, respectively.

%%%%%%%%%%%%%%%%%%%%%%%%%%%%%%%%%%%%
%%  4. Experiments
%%%%%%%%%%%%%%%%%%%%%%%%%%%%%%%%%%%%
\section{Experiments}
\label{sec:exps}

\subsection{Datasets}

We examine the effectiveness of our proposed DPMS on two public SSMIS benchmarks, \textit{i.e.}, the Automated Cardiac Diagnosis Challenge (ACDC) and Left Atrium (LA) datasets.
The ACDC dataset is a 2D benchmark medical dataset focusing on cardiac image analysis, targeting the assessment of cardiac function. It contains 100 Magnetic Resonance Imaging (MRI) scans from 100 patients, which can be divided into a training set containing 70 MRI scans, a validation set containing 10 MRI scans and a testing set containing 20 MRI scans.
The LA dataset is a 3D benchmark medical dataset constructed from the Atrial Segmentation Challenge dataset~\footnote{http://atriaseg2018.cardiacatlas.org/}, which consists of a collection of 100 fully annotated 3D gadolinium-enhanced MRI scans. Following UA-MT~\cite{sss22uamt}, it is divided into a training set containing 80 MRI scans and a validation set containing 20 MRI scans. 

\subsection{Implementation details}
Follow the previous works~\cite{sss22uamt}, we utilize the UNet~\cite{seg15unet} and VNet~\cite{milletari2016v} as our backbones on the ACDC and LA datasets, respectively. We use an SGD optimizer with a momentum of 0.9 and a polynomial learning-rate decay with an initial value of 0.01 to train the student model. 
Following previous studies~\cite{sss22uamt,luo2022semi,wu2022mutual}, training images are randomly cropped into $256 \times 256$ and $112 \times 112 \times 80$ for the 2D ACDC and 3D LA datasets, respectively. We train the segmentation model on the ACDC dataset with a batch size of 24 (12 labeled and 12 unlabeled instances) for 30,000 iterations. On LA, following existing studies, we adopt a batch size of 4 (2 labeled and 2 unlabeled instances) for training 15, 000 iterations. By default, we over-sampling labeled data and set the size ratio $\mu=1$, the momentum parameter $\alpha=0.99$, the maximum loss weight $\lambda_u = 2.0$ for all runs.

{ % table 1: AC DC
\begin{table*}[t]
	\centering
    \begin{threeparttable}
	\resizebox{0.999\textwidth}{!}{
	\begin{tabular}{l|cc|cccc|cc}
		\hline 
		\hline
		\multirow{2}{*}{Method}&\multicolumn{2}{c}{\# Scans used}&\multicolumn{4}{|c}{Metrics}&\multicolumn{2}{|c}{Complexity}\\
		\cline{2-9}
		&Labeled&Unlabeled &Dice(\%)$\uparrow$ &Jaccard(\%)$\uparrow$&95HD(voxel)$\downarrow$&ASD(voxel)$\downarrow$&Para.(M)&MACs(G)\\
		\hline
		U-Net & 3 (5\%) &0 &47.83 &37.01 &31.16 &12.62 &1.81&2.99\\
		U-Net & 7 (10\%) &0 &79.41 &68.11 &9.35 &2.70 &1.81&2.99\\
        U-Net & 14 (20\%) &0 &85.15 &75.48 &6.20 &2.12 &1.81&2.99\\
		U-Net & 70 (All) &0 &91.44 &84.59 &4.30 &0.99 &1.81&2.99\\
		\hline
		UA-MT \cite{sss22uamt} & \multirow{8}{*}{3 (5\%)} &\multirow{8}{*}{67 (95\%)} &46.04 &35.97 &20.08 &7.75 &1.81&2.99\\
		SASSNet \cite{li2020shape}   &  & &57.77 &46.14 &20.05 &6.06 &1.81&3.02\\
		DTC \cite{luo2021semi}   &  &&56.90 &45.67 &23.36 &7.39 &1.81&3.02\\
		URPC \cite{luo2021efficient}  &  & &55.87 &44.64 &13.60 &3.74 &1.83&3.02\\
		MC-Net \cite{wu2022mutual}  & & &62.85 &52.29 &7.62 &2.33 &2.58&5.39\\
        SS-Net \cite{wu2022exploring}  & &  &65.82 &55.38 &6.67 & 2.28 &1.83 &2.99\\
        CT-CT \cite{luo2022semi}   & &  &65.50 & - &16.2 &- &28.93 &2.99 \\
% % % % % % % % % % % % % % %  lb3
        \textbf{DPMS (Ours)} & & &\textbf{88.44} $\pm 0.15$ & \textbf{80.00} $\pm 0.20$ & \textbf{2.03} $\pm 0.56$ & \textbf{0.59} $\pm 0.09$ &1.81 &2.99\\
		\hline
		UA-MT \cite{sss22uamt}& \multirow{8}{*}{7 (10\%)} &\multirow{8}{*}{63 (90\%)} &81.65 &70.64&6.88 &2.02 &1.81&2.99\\
		SASSNet \cite{li2020shape}   &  & &84.50 &74.34 &5.42 &1.86 &1.81&3.02\\
		DTC \cite{luo2021semi}  &  & &84.29 &73.92 &12.81 &4.01 &1.81&3.02\\
		URPC \cite{luo2021efficient} &  & &83.10 &72.41 &4.84 &1.53 &1.83&3.02\\
		MC-Net \cite{wu2022mutual}  &  & &86.44 &77.04 &5.50 &1.84 &2.58&5.39\\
        SS-Net \cite{wu2022exploring} & & &86.78 &77.67 &6.07 &1.40 &1.83&2.99\\
        CT-CT \cite{luo2022semi}   & &  &86.40 & - &8.60 & - &28.93 &2.99\\
% % % % % % % % % % % % % % %  lb7
        \textbf{DPMS (Ours)} & & &\textbf{89.82} $\pm 0.34$ & \textbf{82.06} $\pm 0.51$ & \textbf{1.72} $\pm 0.52$ & \textbf{0.52} $\pm 0.06$ &1.81 &2.99\\
        \hline
		UA-MT \citep{sss22uamt} & \multirow{6}{*}{14 (20\%)} &\multirow{6}{*}{56 (80\%)} &85.87 &76.78 &5.06 &1.54 &1.81&2.99\\
		SASSNet \citep{li2020shape} &  & &87.04 &78.13 &7.84 &2.15&1.81&3.02\\
		DTC \citep{luo2021semi}&  & &86.28 &77.03 &6.14 &2.11 &1.81&3.02\\
		URPC \citep{luo2021efficient} &  & &85.07 &75.61 &6.26 &1.77 &1.83&3.02\\
		MC-Net \citep{wu2022mutual}  &  & &87.83 &79.14 &4.94 &1.52&2.58&5.39\\
% % % % % % % % % % % % % % %  lb14
		\textbf{DPMS (Ours)} & & &\textbf{91.06} $\pm 0.14$ & \textbf{84.03} $\pm 0.23$ & \textbf{1.27} $\pm 0.11$ & \textbf{0.36} $\pm 0.03$ &1.81 &2.99\\
		\hline
		\hline
	\end{tabular}}
    \end{threeparttable}
    \caption{Comparisons with recent SSMIS methods on the ACDC dataset with 3 (5\%), 7 (10\%), 14 (20\%) labels in terms of Dice, Jaccard, 95HD, ASD and model complexities. Results of our proposed DPMS are average over 3 runs, where the mean and standard deviation are reported.}
	\label{tab:result:acdc}
\end{table*}
}

{ % table 2: LA dataset
\begin{table*}[ht]
	\centering
    \begin{threeparttable}
	\resizebox{0.999\textwidth}{!}{
	\begin{tabular}{l|cc|cccc|cc}
		\hline 
		\hline
		\multirow{2}{*}{Method}&\multicolumn{2}{c}{\# Scans used}&\multicolumn{4}{|c}{Metrics}&\multicolumn{2}{|c}{Complexity}\\
		\cline{2-9} 
		&Labeled&Unlabeled &Dice(\%)$\uparrow$ &Jaccard(\%)$\uparrow$&95HD(voxel)$\downarrow$&ASD(voxel)$\downarrow$&Para.(M)&MACs(G)\\
		\hline
        V-Net & 4(5\%) &0 &52.55 &39.60 &47.05 &9.87 &9.44 &47.02\\
		V-Net &8(10\%) &0 &82.74 &71.72 &13.35 &3.26 &9.44 &47.02\\
        V-Net &16(20\%) &0 &86.96 &77.31 &11.85 &3.22 &9.44 &47.02\\
		V-Net &80(All) &0 &91.47 &84.36 &5.48 &1.51 &9.44 &47.02\\
		\hline
		UA-MT \cite{sss22uamt} & \multirow{6}{*}{4 (5\%)} &\multirow{6}{*}{76 (95\%)} &82.26 &70.98 &13.71 &3.82 &9.44 &47.02\\
		SASSNet \cite{li2020shape} &  & &81.60 &69.63 &16.16 &3.58 &9.44 &47.05\\
		DTC \cite{luo2021semi} &  & &81.25 &69.33 &14.90 &3.99 &9.44 &47.05\\
		URPC \cite{luo2021efficient} &  & &82.48 &71.35 &14.65 &3.65 &5.88&69.43\\
		MC-Net \cite{wu2022mutual}	 &  & &83.59 & 72.36 &14.07 &2.70 &12.35 &95.15\\
		SS-Net \cite{wu2022exploring} &  & &86.33 &76.15 &9.97 &2.31 &9.46 &47.17\\
% % % % % % % % % % % % % % %  lb4
        \textbf{DPMS (Ours)} & & &\textbf{89.64} $\pm 0.22$ & \textbf{81.29} $\pm 0.35$ & \textbf{5.99} $\pm 0.40$ & \textbf{1.77} $\pm 0.05$ &9.44 &47.02\\
        \hline
		UA-MT \cite{sss22uamt} & \multirow{6}{*}{8 (10\%)} &\multirow{6}{*}{72 (90\%)} &86.28 &76.11 &18.71 &4.63 &9.44 &47.02\\
		SASSNet \cite{li2020shape} &  & &85.22 &75.09 &11.18 &2.89 &9.44 &47.05\\
		DTC \cite{luo2021semi} &  & &87.51 &78.17 &8.23 &2.36 &9.44 &47.05\\
		URPC \cite{luo2021efficient} &  & &85.01 &74.36 &15.37 &3.96 &5.88&69.43\\
		MC-Net \cite{wu2022mutual} &  & &87.50 &77.98 &11.28 & 2.30 &12.35 &95.15\\
		SS-Net~\cite{wu2022exploring} &  & &88.55 &79.62 &7.49 &1.90 &9.46 &47.17\\
% % % % % % % % % % % % % % %  lb8
        \textbf{DPMS (Ours)} & & &\textbf{90.49} $\pm 0.21$ & \textbf{82.69} $\pm 0.35$ & \textbf{6.39} $\pm 0.20$ & \textbf{1.53} $\pm 0.05$ &9.44 &47.02\\
        \hline
        UA-MT \cite{sss22uamt} & \multirow{6}{*}{16 (20\%)} &\multirow{6}{*}{64 (80\%)} &88.74 &79.94 &8.39 &2.32 &9.44 &47.02\\
		SASSNet \cite{li2020shape} &  & &89.16 &80.60 &8.95 &2.26 &9.44 &47.05\\
        DTC \cite{luo2021semi} &  & &89.52 &81.22 &7.07 &1.96 &9.44 &47.05\\
		URPC \cite{luo2021efficient} &  & &88.74 &79.93 &12.73 &3.66 &5.88&69.43\\
		MC-Net \cite{wu2022mutual} &  & &90.12 &82.12 &8.07 &1.99 &12.35 &95.15\\
% % % % % % % % % % % % % % %  lb16
        \textbf{DPMS (Ours)} & & &\textbf{91.64} $\pm 0.26$ & \textbf{84.62} $\pm 0.43$ & \textbf{5.21} $\pm 0.28$ & \textbf{1.44} $\pm 0.07$ &9.44 &47.02\\
		\hline
		\hline
	\end{tabular}}
    \end{threeparttable}
    \caption{Comparisons with recent SSMIS methods on the 3D LA dataset with 4 (5\%), 8 (10\%), 16 (20\%) labels in terms of Dice, Jaccard, 95HD, ASD and model complexities.}
	\label{tab:result:la}
\end{table*}
}

{ % table 3: Pancreas
\begin{table*}[!htb]
	\centering
    \begin{threeparttable}
	\resizebox{\textwidth}{!}{
	\begin{tabular}{l|cc|cccc|cc}
		\hline 
		\hline
		\multirow{2}{*}{Method}&\multicolumn{2}{c}{\# Scans used}&\multicolumn{4}{|c}{Metrics}&\multicolumn{2}{|c}{Complexity}\\
		\cline{2-9}
		&Labeled&Unlabeled &Dice(\%)$\uparrow$ &Jaccard(\%)$\uparrow$&95HD(voxel)$\downarrow$&ASD(voxel)$\downarrow$&Para.(M)&MACs(G)\\
		\hline
		V-Net &6 (10\%) &0 &54.94 &40.87 &47.48 &17.43 &9.44&41.45\\
		V-Net  & 12 (20\%) &0 &71.52 &57.68 &18.12 &5.41 &9.44&41.45 \\
		V-Net  &62 (All) &0 &82.60 &70.81 &5.61 &1.33 &9.44&41.45\\
		\hline
		UA-MT \citep{sss22uamt} & \multirow{7}{*}{6 (10\%)} &\multirow{7}{*}{56 (90\%)} &66.44 &52.02 &17.04 &3.03 &9.44&41.45\\
		SASSNet \citep{li2020shape} &  & &68.97 &54.29 &18.83 &1.96&9.44 &41.48\\
		DTC \citep{luo2021semi}  &  & &66.58 &51.79 &15.46 &4.16 &9.44 &41.48\\
		URPC \citep{luo2021efficient}  &  & &73.53 & 59.44 &22.57 &7.85 &5.88 &61.21\\
		MC-Net \citep{wu2022mutual}  &  & &69.07 &54.36 &14.53 &2.28 &12.35&83.88\\
		% MC-Net+  (Ours) & & &70.00 &55.66 &16.03 &3.87&9.44&41.45\\
		% \textit{Multi-scale MC-Net+$^*$} & & &\textbf{\textit{74.01}} &\textbf{\textit{60.02}} &\textbf{\textit{12.59}} &\textit{3.34} &5.88 &61.21\\
        \textbf{DPMS (Ours)} & & &\textbf{79.88} $\pm 0.56$ & \textbf{67.14} $\pm 0.48$ & \textbf{7.77} $\pm 1.45$ & \textbf{1.56} $\pm 0.26$ &9.44 &41.45\\
% % % % % % % % % % % % % % %  lb6
		\hline
		UA-MT \citep{sss22uamt} & \multirow{7}{*}{12 (20\%)} &\multirow{7}{*}{50 (80\%)} &76.10 &62.62 &10.84 &2.43 &9.44&41.45\\
		SASSNet \citep{li2020shape}  &  & &76.39 &63.17 &11.06 &1.42&9.44 &41.48\\
		DTC \citep{luo2021semi}  &  & &76.27 &62.82 &8.70 &2.20 &9.44 &41.48\\
		URPC \citep{luo2021efficient}  &  & &80.02 & 67.30 &8.51 &1.98 &5.88 &61.21\\
		MC-Net \citep{wu2022mutual}  &  & &78.17 &65.22 &6.90 &1.55 &12.35&83.88\\
  % % % % % % % % % % % % % % %  lb12
		% MC-Net+  (Ours)  &  & &79.37 &66.83 &8.52 &1.72&9.44&41.45\\
		% \textit{Multi-scale MC-Net+$^*$} & & &\textbf{\textit{80.59}} &\textbf{\textit{68.08}} &\textbf{\textit{6.47}} &\textit{1.74} &5.88 &61.21\\
  \textbf{DPMS (Ours)} & & &\textbf{82.35} $\pm 0.24$ & \textbf{70.38} $\pm 0.30$ & \textbf{5.17} $\pm 0.07$ & \textbf{1.23} $\pm  0.08$ &9.44 &41.45\\
		\hline
		\hline
	\end{tabular}}
    \end{threeparttable}
    % \caption{Comparisons with five state-of-the-art methods on the Pancreas-CT dataset. Note that, the model complexities, \textit{i.e., the number of parameters (Para.) and multiply-accumulate operations (MACs),} are measured during the model inference.}
    \caption{Comparisons with recent SSMIS methods on the 3D Pancreas dataset with 6 (10\%) and 12 (20\%) labels in terms of Dice, Jaccard, 95HD, ASD and model complexities.}
	\label{tab:result:pan}
\end{table*}
}

% \clearpage
\subsection{Comparison with SOTAs}

We compare our DPMS method with recent SSMIS methods, including UA-MT~\cite{sss22uamt}, SASSNet~\cite{li2020shape}, DTC \cite{luo2021semi}, URPC \cite{luo2021efficient}, MC-Net \cite{wu2022mutual}, SS-Net \cite{wu2022exploring} and CT-CT \cite{luo2022semi}. 
We follow the same data partition protocols as UA-MT~\cite{sss22uamt} and SS-Net~\cite{wu2022exploring} to carry on our experiments and report the mean performance averaged over three runs together with the standard error of the mean (SEM). Following previous works~\cite{sss22uamt, wu2022exploring}, we adopt Dice Score (\%), Jaccard Score (\%), 95\% Hausdorff Distance (95HD) in voxel and Average Surface Distance (ASD) in voxel to evaluate the performance of different methods.

We first investigate the effectiveness of our DPMS on the \textbf{2D ACDC dataset} in Table~\ref{tab:result:acdc}. We observe that our DPMS method achieves new state-of-the-art performance under all protocols without introducing any additional parameters or multiply-accumulate operations (MACs) complexity, surpassing the baseline method by a large margin. It should also be noted that when there are only 5\%, 10\% or 20\% labeled data available with the remaining data unlabeled, our DPMS method exceeds the baseline method by over 40\%, 20\% or 3\% in terms of the Dice Score, respectively. Especially when there are only 3 labeled data available, our DPMS method can surpass the previous SOTA SS-Net by over 20\% Dice score.
We can also see that when using 20\% labeled data, our method achieves comparable performance with the fully supervised upper bound (using all labeled data). Surprisingly, DPMS can even achieve better 95HD and ASD results than the fully supervised method, indicating the great potential of our proposed method.
% indicating the robustness of our model against the incorrect annotations.

In Table~\ref{tab:result:la}, we compare our method with recent SSMIS methods on the \textbf{3D LA dataset}. It can be clearly seen from the table that our DPMS can consistently outperform other methods under all partition protocols. When there are 5\%, 10\% and 20\% labeled data available, our method can surpass the previous SOTA method by around 3\%, 2\% and 1\% in terms of the Dice Score, respectively. It should be specially noted that when there are 20\% labeled data available with the rest data unlabeled, our DPMS method can even outperform the fully supervised method in terms of all evaluation metrics, which further indicates the robustness of our method against the potentially incorrect annotations. Thus our method can possibly be a potential candidate to address noisy segmentation problems.

In addition, we further examine our DPMS on 3D Pancreas dataset. With the exactly same parameter settings as on LA dataset, DPMS can clearly outperform other SSMIS methdos on Pancreas. As shown in Table~\ref{tab:result:pan}, DPMS yields a significant improvement of 24.94\% and 10.83\% over the supervised baseline with 10\% and 20\% labeled samples, respectively. With 6 labels, DPMS achieves a higher Dice score of 79.88\% against to the previous SOTA, URPC~\cite{luo2022semi} of 73.53\%. Surprisingly, compared to the fully supervised baseline with all labeled data available, our DPMS can obtain close performance with only 20\% labeled samples.

% \clearpage
%%%%%%%%%%%%%%%%%%%%%%%%%%%%%%%%%%%%
%%  5. Ablation studies
%%%%%%%%%%%%%%%%%%%%%%%%%%%%%%%%%%%%

{ % table: component
\begin{table}[t]
\centering
\begin{threeparttable}
\resizebox{0.95\textwidth}{!}{
\begin{tabular}{cc|ll|ll|ll|ll}
\hline
\hline
\multicolumn{2}{c|}{\textbf{DPMS}}&\multicolumn{2}{c}{\textbf{RV}}&\multicolumn{2}{c}{\textbf{Myo}}&\multicolumn{2}{c}{\textbf{LV}}&\multicolumn{2}{|c}{\textbf{Mean}} \\\cline{1-10} 
Perturbation & Stablization&
Dice(\%)&95HD&
Dice(\%)&95HD&
Dice(\%)&95HD&
Dice(\%)$\uparrow$&95HD$\downarrow$\\ 
\hline
& & 28.54 & 34.20 & 49.51 & 19.97 & 58.59 & 28.87 & 45.55 & 27.68\\
\checkmark & & 55.69 & 14.09 & 44.96 & 18.96 & 55.51 & 16.74 & 52.05 & 16.60\\
& \checkmark& 64.34 & 11.59 & 68.92 & 9.82 & 82.41 & 12.43 & 71.89 & 11.28\\
\checkmark& \checkmark& 86.25 & 1.80 & 87.01 & 1.11 & 92.55 & 1.72 & 88.60 & 1.54\\
\hline
\hline
\end{tabular}
}
\end{threeparttable}
% }
\caption{Ablations on data perturbation and model stabilization when using 3 cases as labeled data on ACDC. RV, Myo, LV represent the right ventricle, myocardium and left ventricle, respectively.}
\label{ablation:dpms:component}
\end{table}
}

{ % table: loss weight
\begin{table}[t]
    \centering
    \begin{tabular}{c|cccccc}
    \toprule
    $\lambda_u$  & 0.5 & 1.0 & 1.5 & 2.0  & 3.0\\
    \midrule
    ACDC (3) & 87.99  & 88.44 & 88.28 & \textbf{88.60} & 88.26   \\
    LA (4) & 88.27  & 89.33 & \textbf{89.72} & 89.64 &  89.35 \\
    \bottomrule
    \end{tabular}
    \caption{Ablations on the loss weight $\lambda_u$, set as $2.0$ by default. }
    \label{tab:dpms:abl:weights}
\end{table}
}

{ % table: threshold
\begin{table}[t]
    \centering
    \begin{tabular}{c|cccccc}
    \toprule
    $\tau$ & 0.7 & 0.75 & 0.8 & 0.85 & 0.9 & 0.95 \\
    \midrule
    ACDC (3)    & 88.21  & 87.78 & 87.97 & 88.07 & 88.42 & \textbf{88.44} \\
    LA (4)    & 88.95 & 89.27 & \textbf{89.33} & 89.14 & 89.03 & 88.87 \\
    \bottomrule
    \end{tabular}
    \caption{Ablations on threshold $\tau$, set as $0.95$ and $0.8$ for dataset ACDC and LA, respectively. }
    \label{tab:dpms:abl:threshold}
\end{table}
}

\subsection{Ablation studies}
\label{exp:abls}

In this section, we analyze the isolated effects of the data perturbation and  model stabilization in DPMS, as well as the sensitiveness of the hyper-parameters used in our method, \textit{i.e.}, the maximum loss weight of the unlabeled data $\lambda_u$ and the pre-defined threshold to select high-confident samples $\tau$. 

We first verify the isolated contributions of the data perturbation and the model stabilization in Table~\ref{ablation:dpms:component}. Here we report the category-wise performance and the mean performance on the ACDC dataset using 3 labeled data (5\% labels) with the remaining data unlabeled, and we evaluate the performance in terms of the Dice Score and 95HD. 
It can be inferred from the table that either the data perturbation or the model stabilization can contribute to the ultimate segmentation performance, where the data perturbation can mainly contribute to the recognition of the hard-to-distinguish classes like the right ventricle (RV), while it shows only a limited contribution to the model on recognizing those easy-to-distinguish classes like the myocardium (Myo) or the left ventricle (LV). The main reason is that data augmentation can provide effective disagreements on unlabeled data, thus contributing to the robustness of the model, while the too-strong data perturbation may potentially destroy the original data distribution, leading to performance degradation.
On the contrary, model stabilization can greatly contribute to the recognition of the model on any class. The main reason is that a stable teacher model, especially the stable BN statistics, can generate more accurate predictions (\textit{i.e.}, pseudo labels) for unlabeled data to train the student model. In this way, the student model could learn useful semantics from the supervision, thus contributing to the model learning.
Equipping the data perturbation with the model stabilization can further contribute to the recognition performance. The main reason is that the stable teacher model can provide accurate supervision for the student model, while the data perturbation provides the prediction disagreement between the student model and the teacher model. Learning from stable predictions generated by the teacher model to mitigate the prediction disagreements will greatly enable the student model to learn useful semantics, and improve the robustness of the student model, which can further contribute to the stability of the teacher model. Therefore, the recognition performance of the model when combining the data perturbation with the model stabilization can greatly improve the SSMIS performance.

We further examine the sensitivity of the hyper-parameters used in our DPMS method in Table~\ref{tab:dpms:abl:weights} and Table\ref{tab:dpms:abl:threshold}. Here we conduct the experiments on both the ACDC and the LA datasets, with only 5\% labeled data available. It can be seen from the tables that our DPMS method is robust to different values of the hyper-parameters, where the variation of the performance using different hyper-parameters is less than 1\% Dice, indicating the robustness of our method. Especially, it can be inferred from Table~\ref{tab:dpms:abl:weights} that a higher weight $\lambda_u$ for the unlabeled data can improve the performance of the model, suggesting that the model will learn more semantics from unlabeled data. In this work, we adopt the $\lambda_u$ as 2.0 on both the ACDC and LA datasets, and we set the threshold $\tau$ as 0.95 and 0.8 for the ACDC and LA, respectively, to obtain the best performance.

\subsection{Qualitative Visualization}
\begin{figure}[th]
  \centering
  \includegraphics[width=0.95\linewidth]{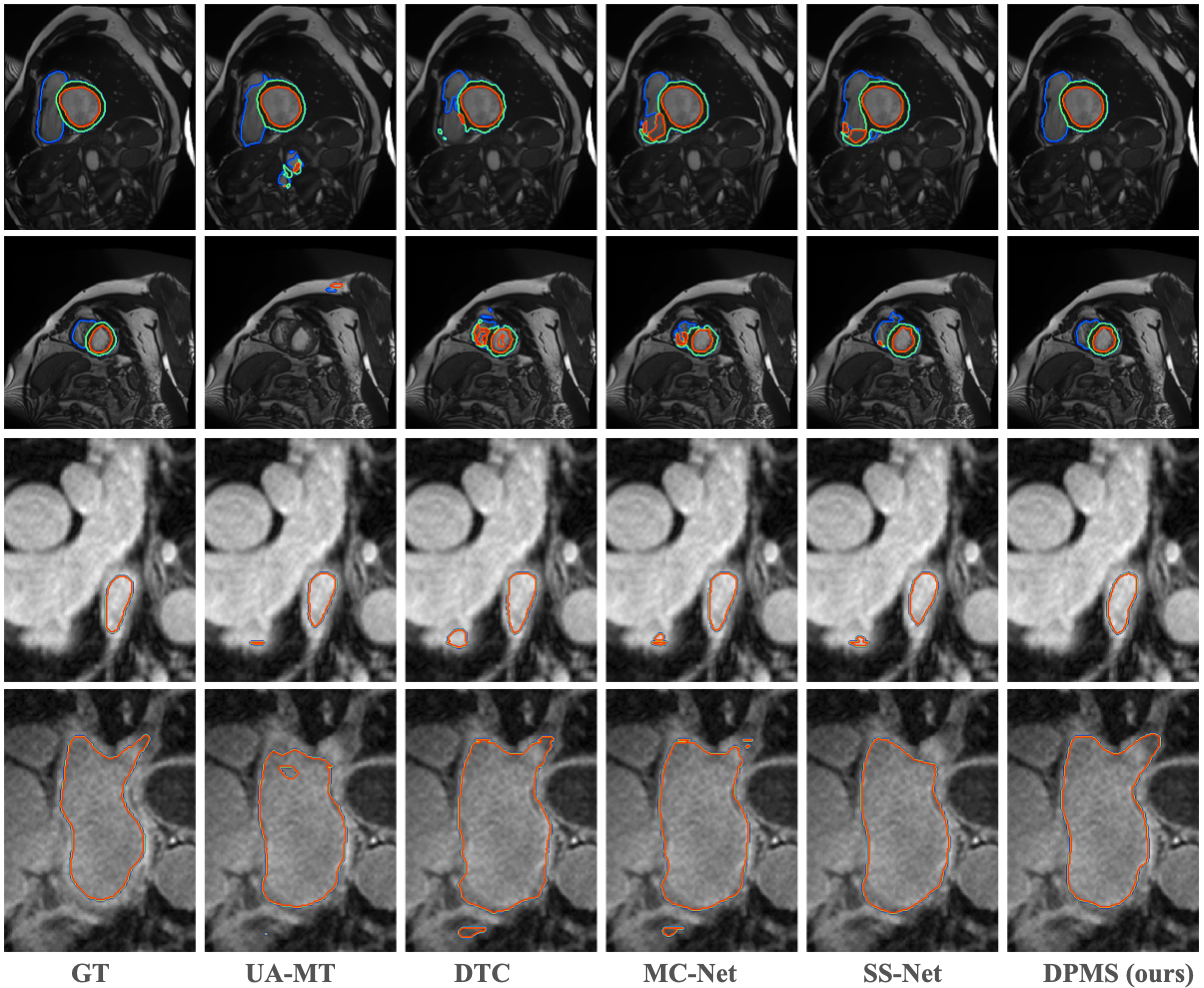}
  \caption{Qualitative results on ACDC (top 2 rows) and LA (bottom 2 rows) using only 5\% labeled data. Columns from left to right denote the segmentation results of the ground-truth, UA-MT~\cite{sss22uamt}, DTC~\cite{luo2021semi}, MC-Net~\cite{wu2022mutual}, SS-Net~\cite{wu2022exploring}, and our proposed DPMS, respectively.
  }
  \label{fig:DPMS:visual}
\end{figure}

Figure~\ref{fig:DPMS:visual} shows some representative qualitative results on ACDC and LA dataset with 5\% labeled data. The UA-MT obtains the worse segmentation results, \eg, not capable of segmenting the myocardium and left ventricle in the second row. Though DTC and MC-Net can obtain better results on ACDC than UA-MT, both methods mis-classify the foreground in the third row on the LA dataset. In contrast, we can observe that many mis-classified regions and ignored segmentation details in the segmentation results of other SSMIS methods  can be successfully corrected and segmented by our proposed DPMS, which further demonstrates the effectiveness of DPMS.

%%%%%%%%%%%%%%%%%%%%%%%%%%%
% 5. Conclusion
%%%%%%%%%%%%%%%%%%%%%%%%%%%
\section{Conclusion}

In this paper, we challenge the prevailing trend observed in recent SSMIS studies, where the focus has shifted towards increasingly complex designs.
Instead, we propose a simple yet highly effective method that emphasizes the significance of data perturbation and model stabilization to boost SSMIS performance. Specifically, we undertake a thorough examination of the essential components of the data, model, and loss supervision in SSMIS. Through in-depth analysis, we find that perturbing and stabilizing strategies play a critical role in achieving promising segmentation performance. 
Experimental results show that DPMS can obtain the new SOTA performance on popular SSMIS benchmarks, especially effective in label-scarce scenarios. We hope our DPMS can serve as a strong baseline and inspire more simple yet effective methods in future SSMIS studies.

% \newpage
%%%%%%%%%%%%%%%%%%%%%%%%%%%
% References
%%%%%%%%%%%%%%%%%%%%%%%%%%%
% \bibliographystyle{plainnat}
\bibliographystyle{unsrt}
\bibliography{refs}

% -------------------------------------------------------
% \newpage
% \appendix
% \section{Submission of papers to NeurIPS 2023}

\end{document}